\def\year{2019}\relax
\documentclass[letterpaper]{article}
\usepackage{aaai19}
\usepackage{times}
\usepackage{helvet}
\usepackage{courier}
\usepackage{url}
\usepackage{graphicx}
\usepackage{amstext}
\usepackage{amsfonts}
\usepackage{enumerate}
\usepackage{subcaption}

\graphicspath{{Figures/}}
\newcommand{\ie}{}\def\ie/{{\em i.e.}}
\newcommand{\eg}{}\def\eg/{{\em e.g.}}
\newcommand{\up}{}\def\up{\textsuperscript}
\newcommand{\lasso}{}\def\lasso/{\textit{LASSO}}
\newcommand{\rscore}{}\def\rscore/{$R_{\text{score}}$}

\title{A Novel Framework for Robustness Analysis of Visual QA Models}
\author{
  Jia-Hong Huang\up{1,2}, Cuong Dao-Duc\up{1*}, Modar Alfadly\up{1*}, Bernard Ghanem\up{1}\\
  \up{1}King Abdullah University of Science and Technology\\
  \up{2}National Taiwan University\\
  \up{*}Authors contributed equally to this work\\
  jiahong.huang@kaust.edu.sa, dao.cuong@kaust.edu.sa, modar.alfadly@kaust.edu.sa, bernard.ghanem@kaust.edu.sa\\
}

\begin{document}
\maketitle
\begin{abstract}
	Deep neural networks have been playing an essential role in many computer vision tasks including Visual Question Answering (VQA). Until recently, the study of their accuracy was the main focus of research but now there is a trend toward assessing the robustness of these models against adversarial attacks by evaluating their tolerance to varying noise levels. In VQA, adversarial attacks can target the image and/or the proposed \emph{main question} and yet there is a lack of proper analysis of the later. In this work, we propose a flexible framework that focuses on the language part of VQA that uses semantically relevant questions, dubbed \emph{basic questions}, acting as controllable noise to evaluate the robustness of VQA models. We hypothesize that the level of noise is positively correlated to the similarity of a basic question to the main question. Hence, to apply noise on any given main question, we rank a pool of basic questions based on their similarity by casting this ranking task as a \lasso/ optimization problem. Then, we propose a novel robustness measure \rscore/ and two large-scale basic question datasets (BQDs) in order to standardize robustness analysis for VQA models.
\end{abstract}

\section{Introduction}

\begin{figure}[t]
	\begin{center}
		\includegraphics{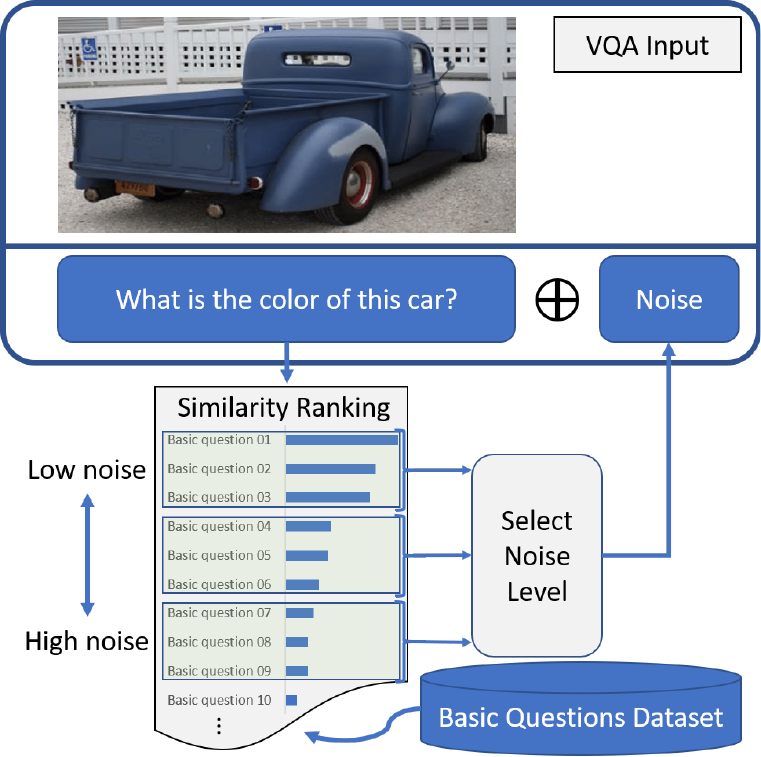}
	\end{center}
	\caption{More robust models tolerate larger noise levels. Thus, To assess the robustness of VQA models, we apply noise at a controllable level to the main question. We sort a dataset of basic questions based on their similarity to the main question and append three basic questions at a time as noise. The robustness is then measured by \rscore/ as the deterioration in accuracy over a given testing dataset.}
	\label{fig:pull_figure}
\end{figure}

Visual Question Answering (VQA) is one of the most challenging computer vision tasks in which an algorithm is given a natural language question about an image and tasked with producing an answer for that image-question pair. Recently, various VQA models \cite{4,9,31,37,41,42,57,58,59} have been proposed to tackle this problem, and their main performance measure is accuracy. In general, any model must have in some sense few aspects of quality relevant to the problem it solves. Accuracy, robustness, stubbornness, and myopia are some examples of such qualities for VQA models that were studied by the community \cite{29}. Obviously, some qualities are more important than others and an overall evaluation of a model should rationalize the interactions of these qualities if possible. In this work, we are interested only in the robustness of VQA models to small \emph{noise} or \emph{perturbation} to the input question, dubbed main question (MQ).

\subsection{Motivation}
The idea of analyzing model robustness as well as training robust models is already a rapidly growing research topic for deep learning models applied to images, \eg/ adversarial attacks in image classification \cite{61,63}. However, most VQA research study other quality aspects that test certain signs of intelligence, such as stubbornness, compositionality, and myopia. Robustness is a concept that despite its importance is heavily misinterpreted and commonly neglected. To the best of our knowledge, we are the first to propose a measure for the robustness of VQA models.

\subsection{Assumptions}
Robustness to some noises (\eg/, motion blur in images) is more important than others (\eg/, salt and pepper) because they are more common. Additionally, some noises have controllable strength, which allows for more sophisticated robustness measures (\eg/, area under the performance curve versus noise level). In this work, we are establishing a framework for analyzing robustness against a type of \emph{controllable additive noise} that is relevant to VQA but is not common. Namely, plain-text concatenation of the MQ with semantically similar questions, dubbed basic questions (BQs); refer to Figure \ref{fig:pull_figure} for a visual illustration. The approach, of considering those BQs as noise, and similar concepts have merit to them and they are studied in-depth in psychology under ``deductive reasoning in human thinking'' despite being a controversial topic \cite{74}. For instance, a person quizzed on the ``color of the bus'' in an image of a car will write down the color of the car while acknowledging the discrepancy. Comparably, if they were asked multiple similar questions, they might be unhinged but answer correctly nevertheless.

\subsection{Robustness Evaluation Framework}
Figure \ref{fig:pull_figure}, depicts the noise generation part of our robustness evaluation. Given a plain-text question (\ie/, the MQ) and a plain-text basic questions dataset (BQD), we start by ranking the BQs in BQD by their similarity to the MQ using some text similarity ranking method; we formulate this ranking as \lasso/ optimization problem. Then, depending on the required level of noise, we take the top, \eg/ $n=3$, ranked BQs and we append them one after the other. The concatenation of these BQs with MQ is the generated noisy question. To measure the robustness of any VQA model, the accuracy over a testing dataset with and without adding this generated noise is compared using our proposed \rscore/.

This framework is posed such that extrapolating to other types of noise is straightforward. For example, \emph{question-words-shuffling noise} can be controlled by the hamming distance (\eg/, the larger the hamming distance the higher level of noise) whereas \emph{question-rephrasing} is a noise that is not trivially controllable. We apply the noise of choice to the questions in a testing dataset, then evaluate the deterioration in accuracy using \rscore/. A more comprehensive robustness evaluation of a VQA model should employ multiple types of noise, possibly ones that could be jointly applied on the image and the MQ at the same time.

\subsection{Contributions}
\begin{enumerate}[i]
	\item We propose a novel framework to measure the robustness of VQA models and test it on six different models.
	\item We propose a new text-based similarity ranking method, \ie/ \lasso/, and compare it against BLEU-1, BLEU-2, BLEU-3, BLEU-4 \cite{49}, ROUGE \cite{68}, CIDEr \cite{67} and METEOR \cite{69}.
	\item We provide the similarity ranking of two large-scale BQDs to test the robustness of VQA models: General Basic Question Dataset (GBQD) and Yes/No Basic Question Dataset (YNBQD).
\end{enumerate}

\section{Related Work}
There is a big effort from the community to tackle the problem of VQA \cite{26,33,54,65,66}. It is a multidisciplinary task that involves natural language progressing (NLP), computer vision and machine learning.

\subsection{Sentence Evaluation Metrics}
Sentence evaluation metrics have been widely used in different areas such as \emph{text summarization} and \emph{machine translation}. In our work, we exploit these metrics to measure the similarity between the MQs and the BQs. BLEU \cite{49} is one of the most popular metrics in machine translation that is based on precision. However, its effectiveness was questioned by \cite{70,71}. METEOR \cite{69}, however, is based on precision and recall. In addition, ROUGE \cite{68} is a popular recall-based metric in the text summarization. It tends to reward longer sentences with higher recall. Moreover, a consensus-based metric, CIDEr \cite{67}, rewards a sentence for being similar to the majority of descriptions written by a human. In our experiments, we take all of the aforementioned metrics to rank BQDs, and our experimental results show that our proposed \lasso/ ranking method achieves better ranking performance.

\subsection{Sentence Embedding}
There exists many method that analyze the relationship between words, phrases and sentences by mapping text into some latent vector spaces \cite{11,43,47}. It was shown that if two phrases share the same context in the corpus, their embedded vectors will be close to each other in the latent space. Skip-thoughts \cite{43} can map text to embedding space using an encoder-decoder architecture of recurrent neural networks (RNNs). The encoder is an RNN with gated recurrent unit (GRU) activation \cite{10}. While the decoder is an RNN with a conditional GRU. We use this model because it performs well on embedding long sentences.

\subsection{Attention Mechanism in VQA}
Attention-based VQA models can attend to local image regions related to the query question \cite{13,15,35,42}. In the pooling step of \cite{42}, the authors use an image-attention mechanism to help determine the relevance of updated question representations to the original. As far as we know, no work has tried to apply the mechanism of language attention to VQA models before \cite{41}. They propose a mechanism of co-attention that jointly performs language and image attention.

\subsection{Multiple Modality Fusion Approaches in VQA}
The VQA task considers features from both the question and image and can be viewed as a multimodal feature fusion task. The authors of \cite{26,57,58,59} have tried to focus on modeling the interactions between two different embedding spaces. In \cite{26}, the authors demonstrate the success of the bilinear interaction between two embedding spaces in deep learning for fine-grained classification and multimodal language modeling. Multimodal Compact Bilinear (MCB) pooling \cite{58} exploits an outer product between visual and textual embedding. Moreover, Multimodal Low-rank Bilinear (MLB) pooling \cite{59} uses a tensor to parametrize the full bilinear interactions between question and image spaces. MCB and MLB are efficiently generalized in \cite{57}.

\subsection{Robustness of Neural Network}
In \cite{61}, the authors analyze the robustness of deep models by adding some perturbations into images and observe how the prediction result is affected. In \cite{72}, the authors mention that VQA models can produce different answers by using slight variations of a query question implying that VQA models do not actually comprehend the asked question.

\section{Methodology}
Our goal is to evaluate the robustness of VQA models over the MQs in the testing set of the popular VQA dataset \cite{4}. For each (image, MQ) pair in the testing set, we concatenate to MQ the most three similar questions (\ie/, BQs) to MQ, obtained by a certain similarity ranking method, over a large-scale basic questions dataset (BQD) as in Figure \ref{fig:pull_figure}. Finally, we compute the accuracy before and after adding this noise and compare them using \rscore/. In this section, we will go through the details of this procedure.

\subsection{Datasets Preparation}
Our BQD is the combination of only unique questions from the training and validation sets of ``real images'' in \emph{the VQA dataset} \cite{4}, which is a total of $186027$ questions. However, because there will be questions in the testing set of $244302$ questions that is also in this BQD, we will exclude them only when executing the similarity ranking.

We will need feature representations of our questions in both the testing set and the BQD. We will use the Skip-thought vector \cite{43} which exploits an RNN encoder with GRU \cite{10} activation to map any English sentence to a feature vector $\mathbf{v} \in \mathbf{R}^{4800}$.

\subsection{Similarity Ranking}
We can project the problem of finding similar questions to a given MQ using this BQD and a similarity ranking method. We will consider two categories of question similarity ranking methods; \emph{direct similarity} and \emph{sparse combination}.

\subsubsection{Direct Similarity} We will need a text similarity (or dissimilarity) scoring method, that takes two sentences and return a similarity (or dissimilarity) score of how ``close'' (or ``far'') are those two semantically from each other. By computing the score of a given MQ to all BQs in the BQD we can directly rank the questions from the most to the least similar. Here, we will use seven scoring methods; BLEU-1, BLEU-2, BLEU-3, BLEU-4, ROUGE, CIDEr and METEOR.

\subsubsection{Sparse Combination} We will need the feature representations of all the questions and a distance measure to determine a sparse combination of BQs needed to represent a given MQ. This can be modelled as follows:
\begin{equation}
	\min_{\mathbf{x}}~\text{distance}\left(\mathbf{A}\mathbf{x}, \mathbf{b} \right)+\lambda \left \| \mathbf{x} \right \|_{1},
	\label{eq:lasso}
\end{equation}
where $\mathbf{b}$ is the feature vector of the MQ, $\mathbf{A}$ is the matrix of feature vectors of all questions in BQD as its columns, and $\lambda$ is a trade-off parameter that controls the sparsity of the solution. Note that all these vectors are normalized to a unit $\ell_2$-norm. To cast this as a \lasso/ optimization problem
$$\text{distance}(\mathbf{q}, \mathbf{p}) = \frac{1}{2}\|\mathbf{q} - \mathbf{p}\|_2^2$$
Different distance metric could give different results especially if they make sense for the latent embedding space. The $i$\up{th} element of the solution $\mathbf{x}$ is the similarity score of the BQ encoded in the $i$\up{th} column of $\mathbf{A}$. We reiterate the importance of making sure that $\mathbf{b}$ is not one of the columns of $\mathbf{A}$. Otherwise, and because we are encouraging sparsity, the ranking will give all other BQs a zero similarity score.

\subsection{Robustness Evaluation}
In the VQA dataset \cite{4}, a predicted answer can be considered partially correct if it matched the answers of less than three human annotators. The overall accuracy is:
\begin{equation}
	\text{Accuracy}_\text{VQA}=\frac{1}{N}\sum_{i=1}^{N}\min\left \{ \frac{\sum_{t\in T_{i}}\mathbb{I}[a_{i}=t]}{3},1 \right \}
\end{equation}
where $\mathbb{I}[\cdot]$ is the indicator function, $N$ is the total number of examples, $a_{i}$ is the predicted answer, and $T_{i}$ is human annotators answer set of the $i^{th}$ image-question pair.

\subsubsection{The \rscore/ of VQA Models}
First, we measure the accuracy of the model on the clean testing set and denote it as $\text{Acc}_\text{VQA}$. Then, we append the top ranked $k$ BQs to each of the MQs and recompute the accuracy of the model on this noisy dataset and we call it $\text{Acc}_\text{BQD}$. Finally, we compute the absolute difference $\text{Acc}_\text{diff} = \left|\text{Acc}_\text{VQA}-\text{Acc}_\text{BQD}\right|$ and report the robustness
\begin{equation}
	R_\text{score} = \text{clamp}_{0}^{1}\left(\frac{\sqrt{m}-\sqrt{{\text{Acc}_\text{diff}}}}{\sqrt{m}-\sqrt{t}}\right)
\end{equation}
Here, we apply this min-max clipping of the score
$$\text{clamp}_{a}^{b}(x) = \max\left(a,{\min\left(b,x\right)}\right)$$
where $0 \leq t < m \leq 1$. The parameters $t$ and $m$ are the tolerance and maximum robustness limit, respectively. In fact, \rscore/ is designed to decrease smoothly from $1$ to $0$ as $\text{Acc}_\text{diff}$ moves from $t$ to $m$ and remains constant outside this range. The rate of change of this transition is exponentially decreasing from exponential to sub-linear in the range $[t, m]$. The reasoning behind this is that we want the score to be sensitive if $\text{Acc}_\text{diff}$ is small, but not before $t$, and less sensitive if it is large, but not after $m$. Note that \rscore/ will be penalized even if $\text{Acc}_\text{VQA}$ was less than $\text{Acc}_\text{BQD}$, \ie/  when we observe an improvement in the accuracy after adding the noise.

\subsection{Basic Questions Datasets}
The size of the BQD has a great impact on the noise generation method. Basically, the more questions we have, the more chance it has to contain similar questions to any given MQ. On the other hand, solving Eq \ref{eq:lasso} for each question in the VQA testing set of $244302$ questions for the $186027$ unique questions in both training and validation set can become computationally expensive. Luckily, we need to do it at most once. We set $\lambda = {10}^{-6}$ and keep the top-$k$ BQs for each MQ produced by solving Eq \ref{eq:lasso}, where $k = 21$. The resultant BQD will have $244302$ instance of this format:
$$\{\text{Image},~MQ,~21~(BQ + \text{corresponding~ similarity~score})\}$$
We keep the same output 10 answers, annotated by different workers through AMT (Amazon Mechanical Turk), for each instance as specified by the VQA dataset \cite{4} in the open-ended and multiple-choice (18 choice) tasks. Approximately $90\%$ of the answers only have a single word and $98\%$ of the answers have no more than three words. In addition, these datasets contain $81434$ testing images from the MS COCO dataset \cite{51}.

Because most of the VQA models have the highest accuracy performance in answering yes/no questions, we have collected two BQDs; Yes/No Basic Question Dataset (YNBQD) and General Basic Question Dataset (GBQD). Now, in practice, our proposed BQDs can be used directly to test the robustness of VQA models without running the \lasso/ ranking method again. In our experiments, we will compare using \lasso/ (\ie/, \emph{sparse combination}) to using other seven \emph{direct similarity} metrics in building BQDs.

\section{Experiments and Analysis}
To analyze our proposed framework, we will perform our experiments on six VQA models; LQI denoting LSTM Q+I \cite{4}, HAV denoting HieCoAtt (Alt, VGG19) and HAR denoting HieCoAtt (Alt, Resnet200) \cite{41}, MU denoting MUTAN without Attention and MUA denoting MUTAN with Attention \cite{57}, and MLB denoting MLB with Attention \cite{4}. On top of that, we will limit ourselves to the open-ended task on the test-dev partition from the 2017\up{th} VQA Challenge \cite{4}, denoted here as dev, unless otherwise specified like using the test-std partition, denoted std.

\subsection{Comparing Similarity Ranking Methods}
In addition to building GBQD and YNBQD using \lasso/, we also generate them using the other similarity metrics (\ie/, BLEU-1, BLEU-2, BLEU-3, BLEU-4, ROUGE, CIDEr and METEOR) as described in the Methodology section. Then, for each MQ, we split the top-21 ranked BQs into seven partition each of which contains three consecutive BQs to control the noise level. More concretely, partition 1, which is (BQ1, BQ2, BQ3), has smaller noise than partition 7, which is (BQ19, BQ20, BQ21), as illustrated in Figure \ref{fig:pull_figure}, with partition 0 being the empty partition. Finally, some VQA models limit the number of words in the MQ, which might lead to trimming the appended noise (\ie/ the BQs in the partition).

\begin{figure}
	\begin{subfigure}[t]{0.49\linewidth}
		\includegraphics{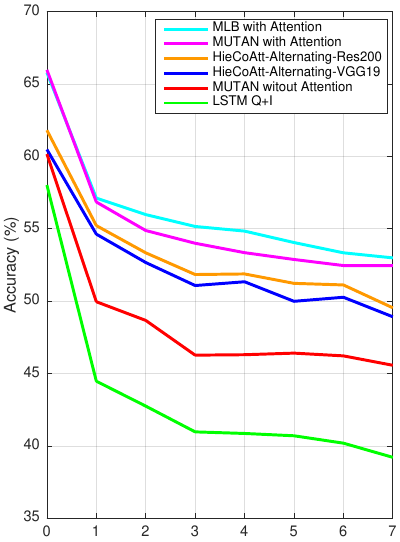}
		\caption{LASSO\_GBQD}
		\label{fig:lasso_gbqd}
	\end{subfigure}
	\hfill
	\begin{subfigure}[t]{0.49\linewidth}
		\includegraphics{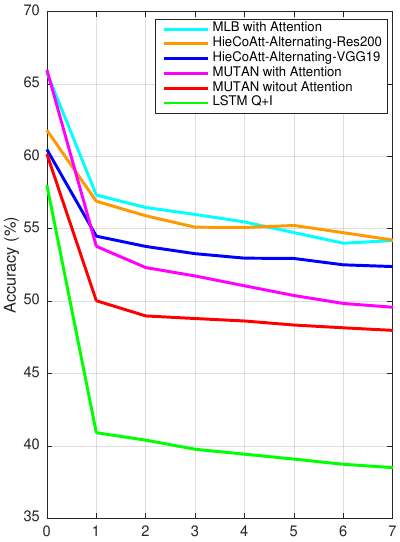}
		\caption{LASSO\_YNBQD}
		\label{fig:lasso_ynbqd}
	\end{subfigure}
	\caption{Compares the accuracy of six VQA models with increasing noise levels from both GBQD and YNBQD. The x-axis is the partition index with 0 meaning MQ without noise. We can see a monotonous trend as the noise increases.}
	\label{fig:lasso}
\end{figure}

\begin{figure*}
	\begin{subfigure}[t]{0.49\linewidth}
		\includegraphics{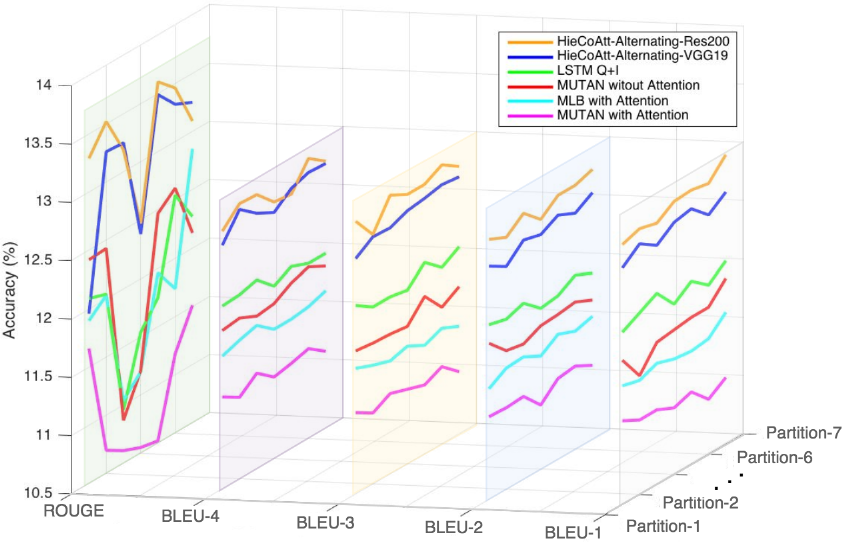}
		\caption{ROUGE, BLEU-4, BLEU-3, BLEU-2 and BLEU-1}
	\end{subfigure}
	\hfill
	\begin{subfigure}[t]{0.49\linewidth}
		\includegraphics{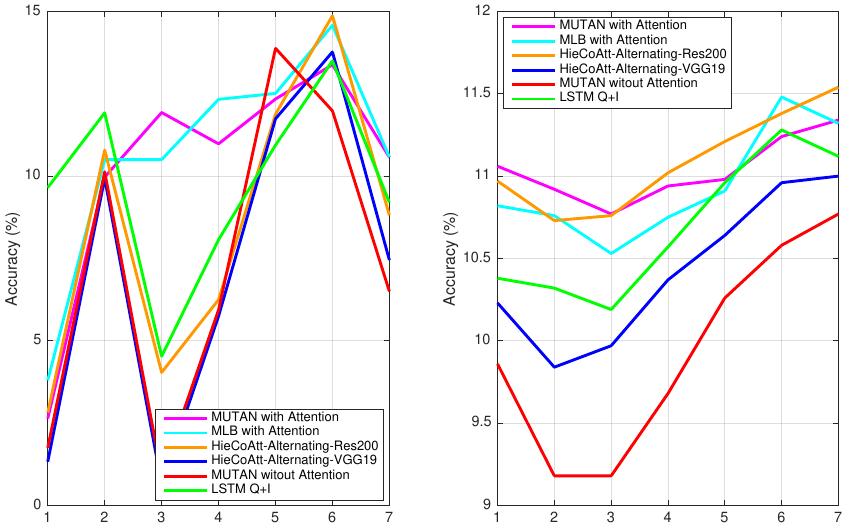}
		\caption{CIDEr~~~~~~~~~~~~~~~~~~~~~~~~~~~~~~~~~~~~~~~~(c) METEOR}
	\end{subfigure}
	\caption{Compares the accuracy of six VQA models with increasing noise levels from GBQD generated by BLEU-1, BLEU-2, BLEU-3, BLEU-4, ROUGE, CIDEr and METEOR. We could not observe a trend as the noise level increase like in Figure \ref{fig:lasso}.}
	\label{fig:others_gbqd}
\end{figure*}

Figure \ref{fig:lasso} shows the accuracy of the six VQA models with the noise coming from GBQD and YNBQD generated by our proposed \lasso/ similarity ranking method. We can see that the accuracy is decreasing as the partition index increases, \ie/ confirming our assumption of treating these BQs as noise and that the noise level is increasing as the similarity score of the BQs decreases. However, we repeated the same experiments on GBQD generated by the other similarity ranking methods in Figure \ref{fig:others_gbqd} and we couldn't see this trend anymore. The plots became less monotonous and acting randomly as we move from partition 1 to partition 7. Also, we observe a big drop in accuracy starting from partition 1 rendering these similarity measures ineffective in this context. In spite of that, various work \cite{1,19,27,28,48} still use them for sentence similarity evaluation because of their simple implementation. This fact signifies the need to develop better similarity metrics, which in turn can be directly evaluated using this flexible framework.

\subsection{Evaluating \rscore/ with \lasso/ Ranking}
\lasso/ Ranking prevails over the other seven similarity metrics. In Table \ref{tbl:lasso_examples}, We present two qualitative \lasso/ ranking examples to showcase the limitations. In Tables \ref{tbl:lasso_gbqd} and \ref{tbl:lasso_ynbqd}, we compare the six VQA models on GBQD and YNBQD and report their \rscore/ on only partition 1 with $t=5\times{10}^{-4}$ and $m=2\times{10}^{-1}$. As we can see, generally speaking, attention-based models (\ie/ HAV, HAR, MUA and MLB) are more robust than non-attentive ones (\ie/ LQI and MU) with HieCoAtt being the most robust. However, MU (non-attentive) is more robust on YNBQD than MUA (the attentive version), while it is not the case on GNBQD. This shows how \rscore/ is strongly tied to the domain of the BQD and that we can only compare models under the same BQD. Note that, we studied only one property of the BQD domain, which is the type of the answer (\ie/, yes-no vs. general).

\begin{table*}
	\begin{subtable}{0.49\linewidth}
		\begin{tabular}{l|l|l}
			Rank & MQ: How old is the car?                 & Score  \\
			\hline
			BQ01 & How old is the truck?                   & 0.2952 \\
			BQ02 & How old is this car?                    & 0.2401 \\
			BQ03 & How old is the vehicle?                 & 0.1416 \\
			BQ04 & What number is the car?                 & 0.1195 \\
			BQ05 & What color is the car?                  & 0.0933 \\
			BQ06 & How old is the bedroom?                 & 0.0630 \\
			BQ07 & What year is the car?                   & 0.0630 \\
			BQ08 & Where is the old car?                   & 0.0368 \\
			BQ09 & How old is the seat?                    & 0.0330 \\
			BQ10 & How old is the cart?                    & 0.0320 \\
			BQ11 & What make is the blue car?              & 0.0281 \\
			BQ12 & How old is the golden retriever?        & 0.0280 \\
			BQ13 & What is beneath the car?                & 0.0239 \\
			BQ14 & Is the car behind him a police car?~~~~ & 0.0223 \\
			BQ15 & How old is the pilot?                   & 0.0198 \\
			BQ16 & How old are you?                        & 0.0172 \\
			BQ17 & How old is the laptop?                  & 0.0159 \\
			BQ18 & How old is the television?              & 0.0157 \\
			BQ19 & What make is the main car?              & 0.0149 \\
			BQ20 & What type and model is the car?         & 0.0148 \\
			BQ21 & What is lifting the car?                & 0.0145 \\
		\end{tabular}
	\end{subtable}
	\begin{subtable}{0.49\linewidth}
		\begin{tabular}{l|l|l}
			Rank & MQ: What is the cat sitting on?         & Score  \\
			\hline
			BQ01 & Where is the cat sitting on?            & 0.2812 \\
			BQ02 & What is this cat sitting on?            & 0.1079 \\
			BQ03 & What is cat sitting on?                 & 0.0548 \\
			BQ04 & What is the cat on the left sitting on? & 0.0526 \\
			BQ05 & What is the giraffe sitting on?         & 0.0499 \\
			BQ06 & What is the cat sitting in the car?     & 0.0469 \\
			BQ07 & That is the black cat sitting on?       & 0.0462 \\
			BQ08 & What is the front cat sitting on?       & 0.0423 \\
			BQ09 & What is the cat perched on?             & 0.0414 \\
			BQ10 & What's the cat sitting on?              & 0.0408 \\
			BQ11 & What is the cat leaning on?             & 0.0370 \\
			BQ12 & What object is the cat sitting on?      & 0.0351 \\
			BQ13 & What is the doll sitting on?            & 0.0289 \\
			BQ14 & How is the cat standing?                & 0.0226 \\
			BQ15 & What is the cat setting on?             & 0.0220 \\
			BQ16 & What is the cat walking on?             & 0.0217 \\
			BQ17 & What is the iPhone sitting on?          & 0.0210 \\
			BQ18 & What is the cat napping on?             & 0.0210 \\
			BQ19 & What is the dog sitting at?             & 0.0201 \\
			BQ20 & What is the birds sitting on?           & 0.0183 \\
			BQ21 & What is the sitting on?                 & 0.0182 \\
		\end{tabular}
	\end{subtable}
	\caption{Demonstrates the limitations of \lasso/ ranking with two examples and their corresponding similarity scores despite performing well in our experiments. From the table, we readily see how the misspellings and sentence re-phrasings can affect the quality of the generated BQD. This can be attributed to the limitation of the encoding and with better semantic encoders and appropriate distance metrics this framework should improve out-of-the-box without any significant changes. In addition, one can utilize the feature vectors to filter out the unique questions to form the BQD instead of relying on raw string comparisons.}
	\label{tbl:lasso_examples}
\end{table*}

\begin{table*}
	\begin{subtable}{0.49\linewidth}
		\begin{tabular}{c | c c c c | c}
			\rscore/ = 0.30  & Other & Num   & Y/N   & All            & Diff           \\ [0.5ex]
			\hline
			dev: partition 1 & 37.78 & 34.93 & 68.20 & \textbf{49.96} & \textbf{10.20} \\
			dev: partition 2 & 37.29 & 35.03 & 65.62 & \textbf{48.67} & \textbf{11.49} \\
			dev: partition 3 & 34.81 & 34.39 & 62.85 & \textbf{46.27} & \textbf{13.89} \\
			dev: partition 4 & 34.25 & 34.29 & 63.60 & \textbf{46.30} & \textbf{13.86} \\
			dev: partition 5 & 33.89 & 34.66 & 64.19 & \textbf{46.41} & \textbf{13.75} \\
			dev: partition 6 & 33.15 & 34.68 & 64.59 & \textbf{46.22} & \textbf{13.94} \\
			dev: partition 7 & 32.80 & 33.99 & 63.59 & \textbf{45.57} & \textbf{14.59} \\
			\hline
			std: partition 1 & 38.24 & 34.54 & 67.55 & \textbf{49.93} & \textbf{10.52} \\
			\hline
			dev: partition 0 & 47.16 & 37.32 & 81.45 & \textbf{60.16} & -              \\
			std: partition 0 & 47.57 & 36.75 & 81.56 & \textbf{60.45} & -              \\
			\hline
		\end{tabular}
		\caption{MUTAN without Attention model evaluation results.}

		\begin{tabular}{c | c c c c | c}
			\rscore/ = \textbf{0.48} & Other & Num   & Y/N   & All            & Diff           \\ [0.5ex]
			\hline
			dev: partition 1         & 44.44 & 37.53 & 71.11 & \textbf{54.63} & \textbf{5.85}  \\
			dev: partition 2         & 42.62 & 36.68 & 68.67 & \textbf{52.67} & \textbf{7.81}  \\
			dev: partition 3         & 41.60 & 35.59 & 66.28 & \textbf{51.08} & \textbf{9.4}   \\
			dev: partition 4         & 41.09 & 35.71 & 67.49 & \textbf{51.34} & \textbf{9.14}  \\
			dev: partition 5         & 39.83 & 35.55 & 65.72 & \textbf{49.99} & \textbf{10.49} \\
			dev: partition 6         & 39.60 & 35.99 & 66.56 & \textbf{50.27} & \textbf{10.21} \\
			dev: partition 7         & 38.33 & 35.47 & 64.89 & \textbf{48.92} & \textbf{11.56} \\
			\hline
			std: partition 1         & 44.77 & 36.08 & 70.67 & \textbf{54.54} & \textbf{5.78}  \\
			\hline
			dev: partition 0         & 49.14 & 38.35 & 79.63 & \textbf{60.48} & -              \\
			std: partition 0         & 49.15 & 36.52 & 79.45 & \textbf{60.32} & -              \\
			\hline
		\end{tabular}
		\caption{HieCoAtt (Alt, VGG19) model evaluation results.}

		\begin{tabular}{c | c c c c | c}
			\rscore/ = 0.36  & Other & Num   & Y/N   & All            & Diff           \\ [0.5ex]
			\hline
			dev: partition 1 & 49.31 & 34.62 & 72.21 & \textbf{57.12} & \textbf{8.67}  \\
			dev: partition 2 & 48.53 & 34.84 & 70.30 & \textbf{55.98} & \textbf{9.81}  \\
			dev: partition 3 & 48.01 & 33.95 & 69.15 & \textbf{55.16} & \textbf{10.63} \\
			dev: partition 4 & 47.20 & 34.02 & 69.31 & \textbf{54.84} & \textbf{10.95} \\
			dev: partition 5 & 45.85 & 34.07 & 68.95 & \textbf{54.05} & \textbf{11.74} \\
			dev: partition 6 & 44.61 & 34.30 & 68.59 & \textbf{53.34} & \textbf{12.45} \\
			dev: partition 7 & 44.71 & 33.84 & 67.76 & \textbf{52.99} & \textbf{12.80} \\
			\hline
			std: partition 1 & 49.07 & 34.13 & 71.96 & \textbf{56.95} & \textbf{8.73}  \\
			\hline
			dev: partition 0 & 57.01 & 37.51 & 83.54 & \textbf{65.79} & -              \\
			std: partition 0 & 56.60 & 36.63 & 83.68 & \textbf{65.68} & -              \\
			\hline
		\end{tabular}
		\caption{MLB with Attention model evaluation results.}
	\end{subtable}
	\begin{subtable}{0.49\linewidth}
		\begin{tabular}{c | c c c c | c}
			\rscore/ = 0.34  & Other & Num   & Y/N   & All            & Diff           \\ [0.5ex]
			\hline
			dev: partition 1 & 51.51 & 35.62 & 68.72 & \textbf{56.85} & \textbf{9.13}  \\
			dev: partition 2 & 49.86 & 34.43 & 66.18 & \textbf{54.88} & \textbf{11.10} \\
			dev: partition 3 & 49.15 & 34.50 & 64.85 & \textbf{54.00} & \textbf{11.98} \\
			dev: partition 4 & 47.96 & 34.26 & 64.72 & \textbf{53.35} & \textbf{12.63} \\
			dev: partition 5 & 47.20 & 33.93 & 64.53 & \textbf{52.88} & \textbf{13.10} \\
			dev: partition 6 & 46.48 & 33.90 & 64.37 & \textbf{52.46} & \textbf{13.52} \\
			dev: partition 7 & 46.88 & 33.13 & 64.10 & \textbf{52.46} & \textbf{13.52} \\
			\hline
			std: partition 1 & 51.34 & 35.22 & 68.32 & \textbf{56.66} & \textbf{9.11}  \\
			\hline
			dev: partition 0 & 56.73 & 38.35 & 84.11 & \textbf{65.98} & -              \\
			std: partition 0 & 56.29 & 37.47 & 84.04 & \textbf{65.77} & -              \\
			\hline
		\end{tabular}
		\caption{MUTAN with Attention model evaluation results.}

		\begin{tabular}{c | c c c c | c}
			\rscore/ = 0.45  & Other & Num   & Y/N   & All            & Diff           \\ [0.5ex]
			\hline
			dev: partition 1 & 46.51 & 36.33 & 70.41 & \textbf{55.22} & \textbf{6.59}  \\
			dev: partition 2 & 45.19 & 36.78 & 67.27 & \textbf{53.34} & \textbf{8.47}  \\
			dev: partition 3 & 43.87 & 36.28 & 65.29 & \textbf{51.84} & \textbf{9.97}  \\
			dev: partition 4 & 43.41 & 36.25 & 65.94 & \textbf{51.88} & \textbf{9.93}  \\
			dev: partition 5 & 42.02 & 35.89 & 66.09 & \textbf{51.23} & \textbf{10.58} \\
			dev: partition 6 & 42.03 & 36.40 & 65.66 & \textbf{51.12} & \textbf{10.69} \\
			dev: partition 7 & 40.68 & 36.08 & 63.49 & \textbf{49.54} & \textbf{12.27} \\
			\hline
			std: partition 1 & 46.77 & 35.22 & 69.66 & \textbf{55.00} & \textbf{7.06}  \\
			\hline
			dev: partition 0 & 51.77 & 38.65 & 79.70 & \textbf{61.81} & -              \\
			std: partition 0 & 51.95 & 38.22 & 79.95 & \textbf{62.06} & -              \\
			\hline
		\end{tabular}
		\caption{HieCoAtt (Alt, Resnet200) model evaluation results.}

		\begin{tabular}{c | c c c c | c}
			\rscore/ = 0.19  & Other & Num   & Y/N   & All            & Diff           \\ [0.5ex]
			\hline
			dev: partition 1 & 29.24 & 33.77 & 65.14 & \textbf{44.47} & \textbf{13.55} \\
			dev: partition 2 & 28.02 & 32.73 & 62.68 & \textbf{42.75} & \textbf{15.27} \\
			dev: partition 3 & 26.32 & 33.10 & 60.22 & \textbf{40.97} & \textbf{17.05} \\
			dev: partition 4 & 25.27 & 31.70 & 61.56 & \textbf{40.86} & \textbf{17.16} \\
			dev: partition 5 & 24.73 & 32.63 & 61.55 & \textbf{40.70} & \textbf{17.32} \\
			dev: partition 6 & 23.90 & 32.14 & 61.42 & \textbf{40.19} & \textbf{17.83} \\
			dev: partition 7 & 22.74 & 31.36 & 60.60 & \textbf{39.21} & \textbf{18.81} \\
			\hline
			std: partition 1 & 29.68 & 33.76 & 65.09 & \textbf{44.70} & \textbf{13.48} \\
			\hline
			dev: partition 0 & 43.40 & 36.46 & 80.87 & \textbf{58.02} & -              \\
			std: partition 0 & 43.90 & 36.67 & 80.38 & \textbf{58.18} & -              \\
			\hline
		\end{tabular}
		\caption{LSTM Q+I model evaluation results.}
	\end{subtable}
	\caption{Compares the accuracy and \rscore/ of six VQA models with increasing noise levels from GBQD generated by \lasso/ evaluated on dev and std. The results are split by the question type; Numerical (Num), Yes/No (Y/N), or Other. There is a multitude of things to consider here in order to give a well-informed interpretation of these scores. One of which is the way $\text{Acc}_\text{diff}$ is defined in Eq \ref{eq:lasso}. The absolute value can be replaced with a ReLU (\ie/, $max(x, 0)$) but this is a design decision which is widely accepted in the literature of adversarial attacks. As we can see, the accuracy of all tested models decrease as they are evaluated on noisier partitions ranked by \lasso/ (i.e., partition 7 has a higher noise level than partition 1).}
	\label{tbl:lasso_gbqd}
\end{table*}

\begin{table*}
	\begin{subtable}{0.49\linewidth}
		\begin{tabular}{c | c c c c | c}
			\rscore/ = 0.30  & Other & Num   & Y/N   & All            & Diff           \\ [0.5ex]
			\hline
			dev: partition 1 & 33.98 & 33.50 & 73.22 & \textbf{49.96} & \textbf{10.13} \\
			dev: partition 2 & 32.44 & 34.47 & 72.22 & \textbf{48.67} & \textbf{11.18} \\
			dev: partition 3 & 32.65 & 33.60 & 71.76 & \textbf{46.27} & \textbf{11.36} \\
			dev: partition 4 & 32.77 & 33.79 & 71.14 & \textbf{46.30} & \textbf{11.53} \\
			dev: partition 5 & 32.46 & 33.51 & 70.90 & \textbf{46.41} & \textbf{11.81} \\
			dev: partition 6 & 33.02 & 33.18 & 69.88 & \textbf{46.22} & \textbf{12.00} \\
			dev: partition 7 & 32.73 & 33.28 & 69.74 & \textbf{45.57} & \textbf{12.18} \\
			\hline
			std: partition 1 & 34.06 & 33.24 & 72.99 & 49.93          & 10.43          \\
			\hline
			dev: partition 0 & 47.16 & 37.32 & 81.45 & \textbf{60.16} & -              \\
			std: partition 0 & 47.57 & 36.75 & 81.56 & \textbf{60.45} & -              \\
			\hline
		\end{tabular}
		\caption{MUTAN without Attention model evaluation results.}

		\begin{tabular}{c | c c c c | c}
			\rscore/ = 0.48  & Other & Num   & Y/N   & All            & Diff          \\ [0.5ex]
			\hline
			dev: partition 1 & 40.80 & 30.34 & 76.92 & \textbf{54.49} & \textbf{5.99} \\
			dev: partition 2 & 39.63 & 30.67 & 76.49 & \textbf{53.78} & \textbf{6.70} \\
			dev: partition 3 & 39.33 & 31.12 & 75.48 & \textbf{53.28} & \textbf{7.20} \\
			dev: partition 4 & 39.31 & 29.78 & 75.12 & \textbf{52.97} & \textbf{7.51} \\
			dev: partition 5 & 39.38 & 29.87 & 74.96 & \textbf{52.95} & \textbf{7.53} \\
			dev: partition 6 & 39.13 & 30.74 & 73.95 & \textbf{52.51} & \textbf{7.97} \\
			dev: partition 7 & 38.90 & 31.14 & 73.80 & \textbf{52.39} & \textbf{8.09} \\
			\hline
			std: partition 1 & 40.88 & 28.82 & 76.67 & 54.37          & 5.95          \\
			\hline
			dev: partition 0 & 49.14 & 38.35 & 79.63 & \textbf{60.48} & -             \\
			std: partition 0 & 49.15 & 36.52 & 79.45 & \textbf{60.32} & -             \\
			\hline
		\end{tabular}
		\caption{HieCoAtt (Alt, VGG19) model evaluation results.}

		\begin{tabular}{c | c c c c | c}
			\rscore/ = 0.37  & Other & Num   & Y/N   & All            & Diff           \\ [0.5ex]
			\hline
			dev: partition 1 & 46.57 & 32.09 & 76.60 & \textbf{57.33} & \textbf{8.46}  \\
			dev: partition 2 & 45.83 & 32.43 & 75.29 & \textbf{56.47} & \textbf{9.32}  \\
			dev: partition 3 & 45.17 & 32.52 & 74.87 & \textbf{55.99} & \textbf{9.80}  \\
			dev: partition 4 & 45.11 & 32.31 & 73.73 & \textbf{55.47} & \textbf{10.32} \\
			dev: partition 5 & 44.35 & 31.95 & 72.93 & \textbf{54.74} & \textbf{11.05} \\
			dev: partition 6 & 43.75 & 31.21 & 72.03 & \textbf{54.00} & \textbf{11.79} \\
			dev: partition 7 & 43.88 & 32.59 & 71.99 & \textbf{54.19} & \textbf{11.60} \\
			\hline
			std: partition 1 & 46.11 & 31.46 & 76.84 & 57.25          & 8.43           \\
			\hline
			dev: partition 0 & 57.01 & 37.51 & 83.54 & \textbf{65.79} & -              \\
			std: partition 0 & 56.60 & 36.63 & 83.68 & \textbf{65.68} & -              \\
			\hline
		\end{tabular}
		\caption{MLB with Attention model evaluation results.}
	\end{subtable}
	\begin{subtable}{0.49\linewidth}
		\begin{tabular}{c | c c c c | c}
			\rscore/ = 0.23  & Other & Num   & Y/N   & All            & Diff           \\ [0.5ex]
			\hline
			dev: partition 1 & 43.96 & 28.90 & 71.89 & \textbf{53.79} & \textbf{12.19} \\
			dev: partition 2 & 42.66 & 28.08 & 70.05 & \textbf{52.32} & \textbf{13.66} \\
			dev: partition 3 & 41.62 & 29.12 & 69.58 & \textbf{51.74} & \textbf{14.24} \\
			dev: partition 4 & 41.53 & 29.30 & 67.96 & \textbf{51.06} & \textbf{14.92} \\
			dev: partition 5 & 40.46 & 27.66 & 68.03 & \textbf{50.39} & \textbf{15.59} \\
			dev: partition 6 & 40.03 & 28.44 & 66.98 & \textbf{49.84} & \textbf{16.14} \\
			dev: partition 7 & 39.11 & 28.41 & 67.44 & \textbf{49.58} & \textbf{16.40} \\
			\hline
			std: partition 1 & 43.55 & 28.70 & 71.76 & 53.63          & 12.14          \\
			\hline
			dev: partition 0 & 56.73 & 38.35 & 84.11 & \textbf{65.98} & -              \\
			std: partition 0 & 56.29 & 37.47 & 84.04 & \textbf{65.77} & -              \\
			\hline
		\end{tabular}
		\caption{MUTAN with Attention model evaluation results.}

		\begin{tabular}{c | c c c c | c}
			\rscore/ = \textbf{0.53} & Other & Num   & Y/N   & All            & Diff          \\ [0.5ex]
			\hline
			dev: partition 1         & 44.42 & 36.39 & 76.94 & \textbf{56.90} & \textbf{4.91} \\
			dev: partition 2         & 43.37 & 34.99 & 76.10 & \textbf{55.90} & \textbf{5.91} \\
			dev: partition 3         & 42.22 & 33.97 & 75.80 & \textbf{55.11} & \textbf{6.70} \\
			dev: partition 4         & 42.52 & 34.21 & 75.33 & \textbf{55.09} & \textbf{6.72} \\
			dev: partition 5         & 42.81 & 34.69 & 75.21 & \textbf{55.23} & \textbf{6.58} \\
			dev: partition 6         & 42.27 & 35.16 & 74.50 & \textbf{54.73} & \textbf{7.08} \\
			dev: partition 7         & 41.95 & 35.14 & 73.64 & \textbf{54.22} & \textbf{7.59} \\
			\hline
			std: partition 1         & 44.93 & 35.59 & 76.82 & 57.10          & 4.96          \\
			\hline
			dev: partition 0         & 51.77 & 38.65 & 79.70 & \textbf{61.81} & -             \\
			std: partition 0         & 51.95 & 38.22 & 79.95 & \textbf{62.06} & -             \\
			\hline
		\end{tabular}
		\caption{HieCoAtt (Alt, Resnet200) model evaluation results.}

		\begin{tabular}{c | c c c c | c}
			\rscore/ = 0.08  & Other & Num   & Y/N   & All            & Diff           \\ [0.5ex]
			\hline
			dev: partition 1 & 20.49 & 25.98 & 68.79 & \textbf{40.91} & \textbf{17.11} \\
			dev: partition 2 & 19.81 & 25.40 & 68.51 & \textbf{40.40} & \textbf{17.62} \\
			dev: partition 3 & 18.58 & 24.95 & 68.53 & \textbf{39.77} & \textbf{18.25} \\
			dev: partition 4 & 18.50 & 24.82 & 67.83 & \textbf{39.43} & \textbf{18.59} \\
			dev: partition 5 & 17.68 & 24.68 & 67.99 & \textbf{39.09} & \textbf{18.93} \\
			dev: partition 6 & 17.29 & 24.03 & 67.76 & \textbf{38.73} & \textbf{19.29} \\
			dev: partition 7 & 16.93 & 24.63 & 67.45 & \textbf{38.50} & \textbf{19.52} \\
			\hline
			std: partition 1 & 20.84 & 26.14 & 68.88 & 41.19          & 16.99          \\
			\hline
			dev: partition 0 & 43.40 & 36.46 & 80.87 & \textbf{58.02} & -              \\
			std: partition 0 & 43.90 & 36.67 & 80.38 & \textbf{58.18} & -              \\
			\hline
		\end{tabular}
		\caption{LSTM Q+I model evaluation results.}
	\end{subtable}
	\caption{Compares the accuracy and \rscore/ of six VQA models with increasing noise levels from YNBQD generated by \lasso/ evaluated on dev and std. The results are split by the question type; Numerical (Num), Yes/No (Y/N), or Other. We notice that the \rscore/ of some models under YNBQD is better than GBQD (\eg/, HAR) and vice versa (\eg/, MUA). So, the \rscore/ can only be compared in different models if they have the same BQD because it can show certain biases in the models under investigation that are related to the type of the BQD. For example, we can clearly see how HAR is more robust towards Yes/No questions than general questions. Whereas, HAV is apparently agnostic to this property (\ie/, the answer type being Yes/No versus general).}
	\label{tbl:lasso_ynbqd}
\end{table*}

\section{Conclusion}
In this work, we propose a novel framework, a semantic similarity ranking method, two large-scale basic question datasets and robustness measure (\rscore/) as a benchmark to help the community build accurate \emph{and} robust VQA models.

\section{Acknowledgments}
This work was supported by the King Abdullah University of Science and Technology (KAUST) Office of Sponsored Research and used the resources of the Supercomputing Laboratory at KAUST in Thuwal, Saudi Arabia.

{\small
\bibliography{references}
\bibliographystyle{aaai}
}
\end{document}